\pdfoutput=1

\documentclass[11pt]{article}

\usepackage[]{ACL2023}

\usepackage{times}
\usepackage{latexsym}

\usepackage[T1]{fontenc}

\usepackage[utf8]{inputenc}

\usepackage{microtype}

\usepackage{inconsolata}

\usepackage{graphicx}
\title{Exploring Hybrid Linguistic Features for Turkish Text Readability}

\author{Ahmet Yavuz Uluslu \\
  Universität Zürich \\
  \texttt{ahmetyavuz.uluslu@uzh.ch} \\\
  \\\And
  Gerold Schneider \\
Universität Zürich \\
  \texttt{gschneid@cl.uzh.ch}\\
  }

\begin{document}
\maketitle
\begin{abstract}
 This paper presents the first comprehensive study on automatic readability assessment of Turkish texts. We combine state-of-the-art neural network models with linguistic features at lexical, morphosyntactic, syntactic and discourse levels to develop an advanced readability tool. We evaluate the effectiveness of traditional readability formulas compared to modern automated methods and identify key linguistic features that determine the readability of Turkish texts.
\end{abstract}

\section{Introduction}
Automatic Readability Assessment (ARA) is an important task in computational linguistics that aims to automatically determine the level of difficulty of understanding a written text, which has implications for various fields, such as healthcare, education, and accessibility \citep{vajjala2021trends}. In the healthcare sector, medical practitioners can use ARA tools to ensure patient information and consent forms are easily understandable \citep{ley1996use}. In the field of education, teachers and learners alike can benefit from ARA systems to adapt materials to the appropriate language proficiency level \citep{kintsch2014reading}. The appropriate readability of technical reports and other business documents is critical to ensure that the intended audience can fully understand the content and can make informed decisions \citep{bushee2018linguistic}. In areas such as cyber-security, readability is particularly important as it can impact response time to risk closures and case materials \citep{smit2021effect}. 

The task of assessing readability presents challenges, particularly when dealing with large corpora of text. Although manual linguistic analysis by domain experts provides valuable insights, it is time-consuming, costly and subject to individual interpretation, which can lead to variability and subjectivity in the annotation results \citep{deutsch2020linguistic}. Recent research in the field has focused on developing automated methods for extracting linguistic predictors and training models for readability assessment.
Despite these crucial applications and developments, the readability efforts in Turkish have largely been confined to traditional readability formulas, such as Flesch-Kincaid \citep{kincaid1975derivation} and its adaptations \citep{atecsman1997turkccede, bezirci2010metinlerin, ccetinkaya2010turkcce}. Several previous studies have pointed out the shortcomings of these formulas \citep{feng2010comparison, feng2009cognitively}. They typically rely on superficial text features such as sentence length and word length. The integration of complex morphological, syntactic, semantic, and discourse features in modern ARA approaches offers the possibility of significantly improving the current readability studies in Turkish. In this paper, we present the first ARA study for Turkish. Our study combines traditional raw text features with lexical, morpho-syntactic, and syntactic information to create an advanced readability assessment tool for Turkish. We demonstrate the effectiveness of our tool on a new corpus of Turkish popular science magazine articles, published for different age groups and educational levels. Our study aims to contribute to the development of automated tools for accessibility, educational research, and language learning in Turkish.

The rest of the paper is organized as follows. In Section 2, we review related work on readability assessment and machine learning-based approaches. In Section 3, we describe our corpus and the linguistic features used in our study. In Section 4, we present the results of our experiments and analyze the effectiveness of our tool. Finally, in Section 5, we conclude our research and discuss future directions.

\section{Previous Work}
The research of quantifying text readability, or the ease with which a text can be read, has a history spanning over a century \citep{dubay2007smart}. Initial research was centered on the creation of lists of difficult words and readability formulas such as Flesch Reading Ease \citep{flesch1948new}, Dale-Chall readability formula \citep{dale1948formula}, Gunning FOG Index \citep{gunning1969fog} and SMOG \citep{mc1969smog}. These formulas are essentially simple weighted linear functions that utilize easily measurable variables such as word and sentence length, as well as the proportion of complex words within a text. Initially developed for the English language, the Flesch Reading Ease formula required recalibration for its application to Turkish, a task undertaken by \citet{atecsman1997turkccede}. However, a significant obstacle in its adoption was Atesman's failure to disclose the statistical variables used in the recalibration process. This gap was later addressed in the work of \citet{ccetinkaya2010turkcce}, which also assigned appropriate grade levels, thus facilitating its practical use in the Turkish educational context. Not long after the adaptation, \citet{bezirci2010metinlerin} introduced an important refinement, akin to the approach taken in the SMOG formula. They propose that features based on polysyllabic words and the total number of syllables present in the document provide distinct indications of text complexity. Accordingly, they included the counts of polysyllabic words (those with 3-, 4-, and 5+ syllables).  \citet{sonmez2003metinlerin} encountered inconsistencies when applying the Gunning FOG Index to Turkish texts which led to the development of their adaptation. The limitations are mainly due to the subjective nature of the formula in identifying complex words and concepts, which contrasts with other formulas that use easier-to-identify criteria such as syllable counts.

Readability assessment has found practical applications in several areas in Turkish, particularly in the fields of medicine and education. For instance, researchers have used the Flesch-Kincaid and Atesman readability formulae to assess the readability of anaesthesia consent forms in Turkish hospitals, which led to valuable insights into how these documents could be optimised for better comprehension \citep{boztas2017readability, boztacs2014evaluating}. In the realm of education, readability studies have been employed to evaluate the complexity of textbooks, thereby ensuring that these crucial learning materials are appropriate for the targeted student age group. For example, research has been conducted to determine the readability levels of Turkish tales in middle-school textbooks, providing insights that could potentially enhance the quality of education by aligning learning materials with students' comprehension abilities \citep{turkben2019readability, tekcsan2020readability, guven2014readability}. While traditional readability formulas have significantly contributed to the field of readability assessment, they are not without their limitations. They often rely heavily on surface-level text features, such as word and sentence length, and fail to account for deeper linguistic and cognitive factors that influence readability \citep{collins2014computational}. 

Readability formulae have inherent limitations that can affect their accuracy and applicability. Given the unique phonetic attributes, sentence formation patterns, and mean syllable length in each language, each language requires its own calibrated readability formula. The validity of studies employing readability formulae calibrated for the English language to evaluate texts in other languages remains questionable. In practice, applying an English-calibrated formula to Turkish texts may result in an overestimation of readability levels. Indeed, most studies that have used this approach have reported inflated levels of readability requirements \citep{akgul2016, akgul2022evaluating} without accounting for the issues of calibration. Furthermore, the evolution of language over time may necessitate periodic re-calibration of these formulas \citep{lee2023traditional}. As language trends evolve and new words and phrases become more common, readability formulas must adapt to remain accurate and relevant. Previous research shows that traditional readability metrics perform unreliably when applied to non-traditional document types such as web pages \citep{petersen2009machine}.

Traditional readability formulas, despite their extensive use, have been criticised for their lack of wide linguistic coverage \citep{feng2009cognitively, feng2010comparison}. These formulas predominantly focus on superficial text features, largely ignoring other linguistic aspects that significantly contribute to text readability. Factors such as syntactic and semantic complexity, discourse structure, and other linguistic branches recognised by \citep{collins2014computational} which are integral to comprehending a text, remain largely unaccounted for in these traditional models. This narrow linguistic focus can lead to inaccuracies in readability assessment, especially when applied to languages or texts with diverse linguistic structures. These scores are relative measures of readability that should be interpreted in the context of the text's overall features and the target audience's reading ability. They are not absolute measures and treating them as such can result in a misunderstanding of the text's actual readability. 

Practitioner errors in applying readability formulas often stem from methodological shortcomings and misinterpretations \citep{wang2013assessing}. The requirement of considerable text sample sizes for traditional measures introduces another impediment, even though the theoretical minimum size for a text sample has yet to be conclusively established. A common methodological error is the inappropriate sampling of text. Some studies might only consider a limited section of a text, such as the first 100 words, leading to skewed results, especially in scientific texts where complexity often increases later in the document. Similarly, the selective assessment of text sections that do not accurately mirror the overall complexity of the text, like focusing solely on the introduction or conclusion, can misrepresent the readability level.

In recent years, research in ARA has shifted from traditional linear models, which use simple metrics such as word and sentence length to estimate the reading level of a text, to fine-grained features \cite{collins2014computational}. These features often include output of machine learning models trained on a combination of word counts, lexical patterns, discourse analysis, morphology, and syntactic structures. There has been an emerging trend toward using neural models for ARA. These models have demonstrated the capacity to implicitly capture the previously mentioned features without the need for manually-defined feature extraction \citep{jawahar2019does}. \citet{martinc2021supervised} and \citet{imperial2021bert} experimented with contextual embeddings of BERT \citep{devlin2018bert} for the readability assessment task, achieving par or better results than feature-based approaches. However, both studies omitted cross-domain evaluation, leading to uncertainty about the extent to which language models rely on topic and genre information, as opposed to readability. Other studies have further explored various strategies to integrate linguistic features with transformer models, promoting a fusion of traditional and neural approaches \citep{lee2021pushing, deutsch2020linguistic}. The state-of-the-art results are currently being achieved by hybrid models that ensemble linguistic features with transformer-based models, highlighting the combined strength of traditional and modern approaches.

\section{Corpus}
Most widely used readability corpora include One Stop English (OSE) \citep{vajjala2018onestopenglish}, the WeeBit corpus \citep{vajjala2012improving} and the Newsela corpus \citep{xu2015problems}. While the majority of these benchmark datasets and corpora are predominantly available in English, there is a growing interest in the development of readability corpora in other languages. In the context of low-resource languages, limited access to digital text resources necessitates reliance on conventional learning materials, such as classroom materials and textbooks. There are currently no existing readability corpora available for Turkish. 

\subsection{TUBITAK PopSci Magazine Readability Corpus}
Our corpus was constructed using popular science articles from TUBITAK Popular Science Magazines \footnote{\url{https://yayinlar.tubitak.gov.tr/}} spanning the period 2007 to 2022. The articles are openly published and made available for non-commercial redistribution and research purposes. We selected 2250 articles from three magazines, each catering to readers of different age groups. These magazines include Meraklı Minik (for ages 0-6), Bilim Çocuk (for ages 7+), and Bilim ve Teknik (for ages 15+). Accordingly, we consider the articles from these magazines as elementary, intermediate, and advanced level reading material. Our corpus is non-parallel and encompasses a diverse range of topics, including instructions for laboratory experiments and brief articles about recent scientific discoveries. This characteristic is similar to that of the WeeBit corpus \citep{vajjala2012improving}, which also includes articles from various topics and resources. Given that the articles in our corpus are written by experts and specifically tailored for distinct age groups, it can be appropriately regarded as an 'expert-annotated' corpus. We used a off-the-shelf pdf-to-text converter to extract the relevant article text and manually corrected the articles to ensure the conversion accuracy of Turkish characters and the layout integrity. Table~\ref{tab:descriptive-stats} displays descriptive statistics for the finalized corpus. 

\begin{table}[ht]
\centering
\footnotesize
\begin{tabular}{|c|c|c|c|}
\hline
\textbf{Level} & \textbf{Avg. Words} & \textbf{Std. Dev} & \textbf{Nr. of Articles} \\
\hline
ELE & 120.95 & 67.35 & 750 \\
\hline
INT & 154.99 & 93.57 & 750 \\
\hline
ADV & 327.08 & 187.54 & 750 \\
\hline
\end{tabular}
\caption{Descriptive Corpus Statistics}
\label{tab:descriptive-stats}
\end{table}

As expected, the advanced texts display a greater average length compared to the elementary texts. However, the high standard deviation values for each level indicate that other factors beyond text length may have a significant impact on determining the reading level of a given text.

We also performed a preliminary analysis on the three reading levels of the corpus using traditional formulae and showed the results in Table~\ref{tab:readability-stats}, presenting readability metrics Atesman, Cetinkaya-Uzun and Type-Token Ratio (TTR). Atesman and Cetinkaya readability scores decrease from one level to the next indicating that texts become more complex at higher reading levels. In contrast, the TTR score increases suggesting that texts become more diverse and less repetitive at higher reading levels. It should also be noted that the readability levels of the elementary-level articles in both formulas were not suitable for the intended age group and that the magazine's disclaimer states that certain articles may require the assistance of an adult or parent. Table~\ref{tab:example_sentences} presents examples of articles representing each of the three reading levels.

\begin{table}[h]
    \centering
    \begin{tabular}{|c|c|c|c|}
        \hline
        \textbf{Feature} & \textbf{ELE} & \textbf{INT} & \textbf{ADV} \\ \hline
        Atesman & 66.06 & 59.73 & 42.32 \\ \hline
        Cetinkaya & 39.31 & 36.62 & 29.81 \\ \hline
        TTR & 0.65 & 0.71 & 0.76 \\ \hline
    \end{tabular}
    \caption{Readability features across reading levels}
    \label{tab:readability-stats}
\end{table}

\begin{table*}[htbp]
\centering
\begin{tabular}{|c|p{\dimexpr\linewidth-2\tabcolsep-2.8cm\relax}|}
\hline
\textbf{Reading Level} & \textbf{Example} \\ \hline
Elementary & Burası bir doğa koruma merkezi. Burada annesi ve babası olmayan turna yavruları var. Merkezde çalışanlardan biri özel bir giysi giyip koluna bir turna kuklası geçirmiş. \textit{(This is a nature conservation centre. There are crane chicks without a mother and father. One of the workers at the centre wears a special suit and a crane puppet on his arm.)} \\ \hline
Intermediate & Robotlar, insanların yaptığı işleri, onların yerine yapan karmaşık makinelerdir. Bu işleri yapmak için programlanırlar. Otomatik olarak ya da uzaktan kumanda edilerek belirli komutları yerine getirirler. \textit{(Robots are complex machines that do the jobs that humans do, instead of them. They are programmed to do these jobs. They fulfil certain commands automatically or by remote control.)} \\ \hline
Advanced & Pek çok canlıda manyetik algının varlığı bilimsel olarak biliniyor. Bakteri, salyangoz, kurbağa ve ıstakoz gibi canlılar Dünya’nın manyetik alanını algılıyor, göçmen kuşlar ve deniz kaplumbağaları yönlerini bu sayede buluyor, köpekler eğitildiklerinde saklanmış çubuk mıknatısın yerini gösterebiliyor. \textit{(The existence of magnetic perception in many living things is scientifically known. Bacteria, snails, frogs and lobsters can sense the Earth's magnetic field, migratory birds and sea turtles can navigate, and dogs can point out a hidden bar magnet when trained to do so.)} \\ \hline
\end{tabular}
\caption{Example sentences for three reading levels}
\label{tab:example_sentences}
\end{table*}

\section{Linguistic Features}
In this study, we explore five subgroups of linguistic features from our Turkish readability corpus: traditional or surface-based features, syntactic features, lexico-semantic features, morphological features, and discourse features. We employ spaCy v3.4.0 \citep{honnibal2020spacy} with the pre-trained tr\_core\_news\_trf model\footnote{\url{https://huggingface.co/turkish-nlp-suite/tr_core_news_trf}} for the majority of general tasks, including entity recognition, POS tagging, and dependency parsing.  We use the Stanford Stanza parser version 1.5.0 \citep{qi2020stanza} for constituency parsing.

\subsection{Traditional Features (TRAD)}
Traditional or surface-based features are commonly used to predict the readability of Turkish texts, and we also adopt them as a baseline for our study. Specifically, we extract 7 traditional features, including Turkish adaptations of well-known readability formulas such as Atesman and Cetinkaya-Uzun, as well as average numbers of words and syllables per document. As noted by \cite{bezirci2010metinlerin} in their evaluation of the Turkish readability formulae, the impact of the number of polysyllabic words on text complexity is different from that of the total number of syllables present in the text. Therefore, we also included the counts of polysyllabic words (3-, 4-, and 5+ syllables) as separate features in our analysis. 

\subsection{Syntactic Features (SYNX)}
Syntactic properties have a significant impact on the overall complexity of a given text, which serves as an important indicator of readability. We extract an array of syntactic features that capture various dimensions of sentence structure.

\textbf{Phrasal and dependency type features:} Reading abilities are related to the ratios involving clauses in a text \citep{lu2010automatic}. We extract features based on noun and verb phrases at sentence and article levels. We integrate features based on the unconditional probabilities of their dependency-based equivalents \citep{dell2011read}. These encompass various types of syntactic dependencies, including subject, direct object, and modifier, among others.

\textbf{Parse tree depth features:} The depth and structure of dependency trees in a text can reflect the level of sentence complexity. Following this principle, we extract the average and maximum depths of the constituency and dependency tree structures present in the text \citep{dell2011read}.

\textbf{Part-of-Speech features:} Part-of-speech (POS) tags provide essential information about the syntactic function of words in sentences. Adapting the work of \citet{tonelli2012making} and \citet{lee2021pushing}, we include features based on universal POS tag counts. Such features offer insights into the distribution and usage of different word categories, adding another layer of syntactic information. 

\subsection{Lexico-Semantic Features (LXSM)}
Lexico-semantic features are a set of linguistic attributes that can reveal the complexity of a text's vocabulary. These features can be used to identify specific words or phrases that may pose difficulty or unfamiliarity to readers \citep{collins2014computational}.

\textbf{Lexical Variation features:} Secondary language acquisition research has found a correlation between the diversity of words within the same Part-Of-Speech (POS) category and the lexical richness of a text \cite{vajjala2012improving}. We extract noun, verb, adjective, and adverb variations, which represent the proportion of the respective category’s words to the total.

\textbf{Type Token Ratio (TTR) features:}
TTR is a commonly used metric to quantify lexical richness and has been widely employed in readability assessment studies. We compute five distinct variations of TTR from \citep{vajjala2012improving}. The standard TTR variations of a text sample are susceptible to the text length, which can introduce bias in the readability assessment. To address this limitation, we also consider the Moving-Average Type–Token Ratio (MATTR) \citep{covington2010cutting}. The MATTR mitigates the length-dependency issue by calculating the TTR score within a moving window across the text. 

\textbf{Psycholinguistic features:}
We adopted word frequencies obtained from the Turkish psycholinguistic database created by \citet{acar2016turkish}. This resource was built from transcriptions of children's speech and corpora of children's literature, thus containing words commonly acquired during early development. It also includes words typically acquired during adulthood from a standard corpus. We extracted the average word and sentence frequency for both early and late-acquired words. We calculate features based on the average log10 values similar to the SubtlexUS corpus \citep{brysbaert2009moving}.

\textbf{Word Familiarity features:} Familiarity with specific words can greatly affect readability. Based on prior work on Italian \citep{dell2011read} and French \citep{franccois2012ai} readability studies, we assessed the vocabulary composition of the articles using a reference list of 1700 basic words essential for achieving elementary reading proficiency in Turkish. This list, a combination of the first 1200 words taught to children aged 0-6 \citep{keklik2010turkccede} and a set of essential words from an open-access textbook\footnote{\url{https://www.turkishtextbook.com/most-common-words/}} for learning Turkish, provides a benchmark for vocabulary familiarity. We calculated the percentage of unique words (types) in the text based on this reference list, performed on a lemma basis.

\subsection{Morphological features (MORPH)}
Morphological complexity plays a significant role in readability assessment, particularly in languages that are morphologically richer than English such as German \cite{hancke2012readability} and Basque \cite{gonzalez2014simple}. In our study, we integrate the Morphological Complexity Index (MCI) from \citet{brezina2019morphological}. The MCI captures the variability of morphological exponents of specific parts-of-speech within a text by comparing word forms with their stems. We calculate MCI features for verbs, nouns, and adjectives, considering different sample sizes and sampling techniques with and without repetition. 
MCI has been leveraged in cross-lingual readability assessment frameworks, proving its applicability across languages with varying morphological structures \citep{weiss2021using}. However, these studies have not explored agglutinative languages such as Turkish and Hungarian. 

\subsection{Discourse features (DISCO)}
The final group of features we examine are entity density features. The presence and frequency of entities within a text can significantly impact the cognitive load required for comprehension. Entities often introduce new conceptual information, thereby increasing the burden on the reader's working memory. This relationship between entities and readability was previously shown by \citet{feng2009cognitively,feng2010comparison}.

\section{Experiments}
We experiment with four different setups: trad-baseline (non-neural model with shallow features), modern-baseline (non-neural model with linguistic features), neural (pretrained transformer models), and hybrid (modern-baseline + neural). We use 10-fold cross-validation (10FCV) and evaluate our models using standard metrics such as accuracy, precision, recall, and macro F1-score. Specifically, we choose traditional learning algorithms such as Logistic Regression, Support Vector Machines, Random Forest and XGBoost as our baseline models. We perform a randomised search to explore a reasonable range of hyper-parameter values. We apply a grid search to identify the optimal combination of hyper-parameter values within this range. 

\subsection{Non-Neural Models with Linguistic Features}
Given the lack of available baselines for the readability task in Turkish, our first objective is to establish a baseline for the readability task. This baseline (trad-baseline) is designed to be on par with traditional readability formulas and is reliant on shallow linguistic features such as sentence and word lengths. By establishing this baseline, we are effectively creating a benchmark that allows for meaningful comparison between the traditional readability formulas, which are the only available methods in readability assessment for Turkish. We expand our feature set and include a more diverse set of linguistic feature groups (modern-baseline). We are interested in the performance of individual features, but we also aim to identify the best-performing combinations when these features are assembled into linguistic groups.

\subsection{Neural Models}
We extend the established usage of transformer-based models in readability assessment \citep{deutsch2020linguistic, martinc2021supervised, lee2021pushing} and opt for the BERTurk model\footnote{\url{https://huggingface.co/dbmdz/bert-base-turkish-uncased}} for our analysis. We tested multiple learning rates and batch sizes to ascertain the optimal configuration for our task. Specifically, we examined the learning rates of [1e-5, 2e-5, 3e-5, 1e-4] and the batch sizes of [8, 16, 32]. Our final model used AdamW optimizer, linear scheduler with 10\% warmup steps, batch size of 8, and learning rate of 3e-5. The sequence lengths of our input documents were all set to 512 tokens. We fine-tune our model for three epochs. 

\subsection{Hybrid Model}
In our study, we experiment with a hybrid model approach that aims to leverage the strengths of both neural and non-neural models in an ensemble learning strategy. The premise behind the hybrid model is based on the observation that while neural models such as BERT have demonstrated robust performance across diverse tasks, they could still benefit from incorporating human-defined linguistic features, which have been key components in traditional non-neural models \citep{deutsch2020linguistic}. Our hybrid model takes a straightforward approach similar to that of \citet{imperial2021bert} and \citet{lee2021pushing}. It combines the soft label predictions generated by the neural model with handcrafted features. These features are then used as input to a non-neural (Random Forest) model. 

\begin{table}[ht]
\centering
\begin{tabular}{|l|c|c|c|c|}
\hline
\textbf{Model} & \textbf{Acc (\%)} & \textbf{Rec} & \textbf{Prec} & \textbf{F1} \\
\hline
SVM & 78.1 & 78.1 & 79.0 & 77.6 \\
\hline
\textbf{RandomF} & \textbf{85.3} & \textbf{85.3} & \textbf{85.1} & \textbf{85.1} \\
\hline
LogR & 83.7 & 83.6 & 83.7 & 83.5 \\
\hline
XGBoost & 84.1 & 84.0 & 84.0 & 83.7 \\
\hline
\end{tabular}
\caption{Performance comparison (modern-baseline) of readability models}
\label{tab:model_performance}
\end{table}

\section{Results}
We compare the performance of traditional and modern baselines to illustrate the process of arriving at the best-performing model. The process of feature and model selection for the baseline models was carried out based on the results obtained from different combinations.

\subsection{Baseline: Feature and Model Evaluation}

\begin{table}[ht]
\centering
\begin{tabular}{|l|c|}
\hline
\textbf{Linguistic Features} & \textbf{Acc (\%)} \\
\hline
TRAD & 65.7 \\
\hline
+ LXSM & 76.4 \\
\hline
+ SYN & 82.5 \\
\hline
+ MORPH & 83.6 \\
\hline
+ DISCO \textbf{(ALL)} & \textbf{85.3} \\
\hline
\end{tabular}
\caption{Incremental contribution of each feature to the RandomF model}
\label{tab:features}
\end{table}

Through evaluation of four distinct models, namely Support Vector Machines (SVM), Random Forest (RandomF), Logistic Regression (LogR), and XGBoost, we assessed combinations of five different linguistic groups: traditional (TRAD), lexico-semantic (LXSM), syntactic (SYNX), morphological (MORPH), and discourse (DISCO) features. Table \ref{tab:model_performance} provides a comparative view of these models' performance when trained using the full combination. Among the four models evaluated, the Random Forest model delivered the highest performance with  85.3\%. Importantly, all of the linguistic groups used provide orthogonal or distinct information. Table \ref{tab:features} demonstrates how each contributing linguistic group incrementally improves the accuracy of the Random Forest model. Their combined strength ultimately achieves the highest overall accuracy score.

\begin{table}[ht]
\centering
\begin{tabular}{|l|c|c|c|c|}
\hline
\textbf{Model} & \textbf{Acc} & \textbf{Prec} & \textbf{Rec} & \textbf{F1} \\
\hline
trad-baseline & 65.7 & 67.5 & 66.8 & 66.7 \\
\hline
modern-baseline & 85.3 & 85.3 & 85.1 & 85.1 \\
\hline
neural & 92.8 & 93.1 & 92.6 & 92.8 \\
\hline
\textbf{hybrid} & \textbf{96.1} & \textbf{96.1} & \textbf{95.6} & \textbf{95.8} \\
\hline
\end{tabular}
\caption{Performance comparison of readability approaches}
\label{tab:model_performance}
\end{table}

The varying levels of performance between different approaches is demonstrated in Table \ref{tab:model_performance}. The hybrid model, which combines the strengths of both traditional and neural methodologies, outperforms all other models, securing the highest values for accuracy, precision, recall, and F1 score. Following the hybrid model, the neural model performs the best. The neural model (BERT) demonstrates an enhanced ability to capture nuanced characteristics of text readability, exhibiting superior performance to the baseline models without any handcrafted linguistic features. The modern baseline, incorporating five different linguistic subgroups, achieves superior performance compared to the traditional baseline. This highlights the advantage of leveraging an extended set of linguistic features over merely relying on surface-level features typical of traditional readability formulae.

\section{Discussion}
\subsection{Model Interpretation}

In order to gain insights into the significance of individual linguistic features within our best-performing model, the RF model, we utilised two well-established model interpretation techniques specifically designed for Random Forest models: Feature Permutation and Mean Decrease in Impurity (MDI) as shown in Figure 1 and 2.

\begin{figure}[htbp]
  \centering
  \includegraphics[width=0.74\linewidth]{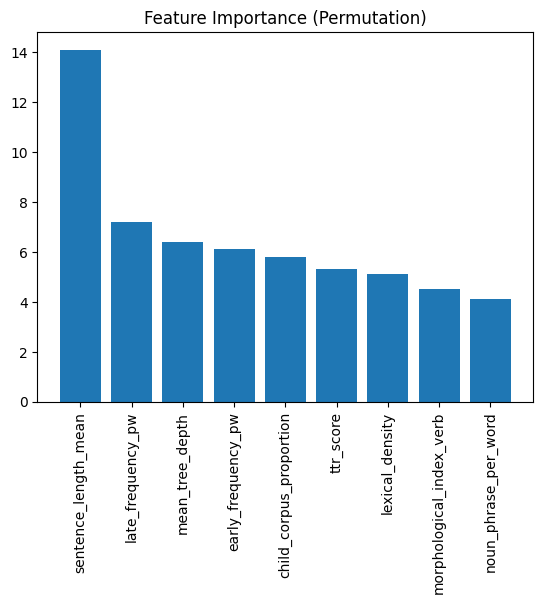}
  \caption{Feature importance by permutation on full model}
  \label{fig:confusion_matrix}
\end{figure}

\begin{figure}[htbp]
  \centering
  \includegraphics[width=0.74\linewidth]{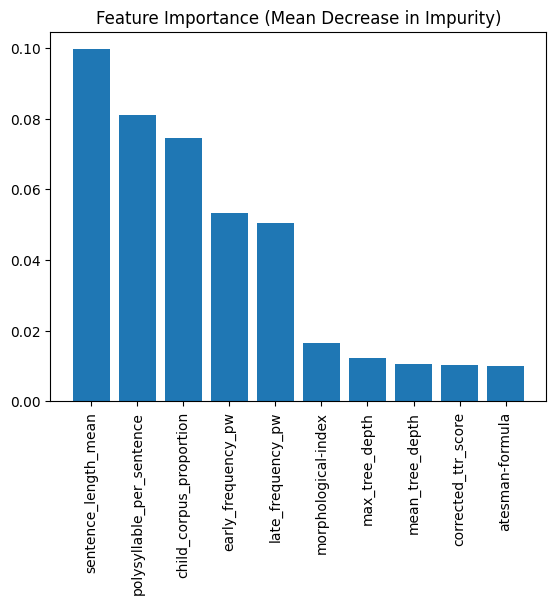}
  \caption{Feature Importance importance by MDI on full model}
  \label{fig:confusion_matrix}
\end{figure}

\subsection{Feature Correlation}

We also considered model-independent analysis through Spearman correlation to gain additional perspective into the importance of features with respect to readability levels. Table~\ref{table:correlation_features} presents the ten features with the highest Spearman correlation coefficients highlighting the significance of readability assessment.

\begin{table}[ht]
\centering
\begin{tabular}{|l|c|c|}
\hline
\textbf{Group} & \textbf{Feature} & \boldmath{$\rho$} \\
\hline
TRAD & Sentence Length Mean & 0.487 \\
\hline
TRAD & Polysyllable Count & 0.467 \\
\hline
LXSM & Child Corpus Proportion & 0.433 \\
\hline
SYNX & Mean Tree Depth & 0.419 \\
\hline
LXSM & Lexical Verb Variation & 0.403 \\
\hline
LXSM & Early Frequency PW & 0.385 \\
\hline
LXSM & Corrected TTR Score & 0.352 \\
\hline
LXSM & Lexical Density & 0.321 \\
\hline
LXSM & Lexical Noun Variation & 0.297 \\
\hline
SYNX & Noun Phrase Per Word & 0.278 \\
\hline
\end{tabular}
\caption{Top ten features ranked by their Spearman correlation coefficients}
\label{table:correlation_features}
\end{table}

\subsection{Lingustic Features}
The analysis of feature importance consistently highlights the significant role of simple measures such as average sentence length and polysyllable counts. These findings align with previous research, where it has been shown that even compared to more complex feature extraction methods, a simple measure such as sentence length can indirectly capture multiple linguistic aspects of readability. Furthermore, our analysis demonstrates that lexico-semantic features play a prominent role in determining readability. This is evident from the performance improvement observed when including LXSM linguistic feature set in the modern-baseline method. It indicates that while traditional features are indeed valuable, incorporating fine-grained information at the semantic and lexical level can lead to an even better understanding of overall readability. The consistent presence of the syntactic feature "mean tree depth" further supports the relationship between sentence length and syntactic complexity. The correlation between mean tree depth and mean sentence length suggests that the structural complexity captured by syntactic features aligns with the overall complexity of sentences.

\section{Conclusion}
We introduced a new readability corpus based on popular science magazine articles, providing a valuable resource for future research in Turkish readability assessment. By exploring the effectiveness of linguistic features at different levels, we have demonstrated their superiority over traditional readability formulae and shallow-level features. Our findings emphasise the importance of incorporating fine-grained linguistic features, as they provide more comprehensive insights into the complexity of Turkish texts. We showed the potential of hybrid models that combine fine-grained features with neural models by leveraging the strengths of both linguistic features and state-of-the-art transformers.

\section*{Acknowledgements}
We would like to pay special thanks to Sefa Kalkan from Charles University for his contributions to the construction of the readability corpus.
\bibliography{anthology,custom}
\bibliographystyle{acl_natbib}

\end{document}